\documentclass[letterpaper]{article} 
\usepackage{aaai24}  
\usepackage{times}  
\usepackage{helvet}  
\usepackage{courier}  
\usepackage[hyphens]{url}  
\usepackage{graphicx} 
\usepackage{natbib}  
\usepackage{caption} 
\usepackage{algorithm}
\usepackage{algorithmic}
\usepackage{tabularx,ragged2e}
\usepackage{booktabs}
\usepackage{color}
\usepackage{amsfonts}
\usepackage{amssymb}
\usepackage{amsmath}
\usepackage{newfloat}
\usepackage{listings}
\DeclareCaptionStyle{ruled}{labelfont=normalfont,labelsep=colon,strut=off} 
\floatstyle{ruled}
\newfloat{listing}{tb}{lst}{}
\floatname{listing}{Listing}
\newcommand{\addition}[1]{\textcolor{black}{#1}}
\newcommand{\revise}[1]{\textcolor{black}{#1}}

\usepackage{hyperref} 
\hypersetup{
    colorlinks=true,
    linkcolor=black,
    citecolor=black,
    urlcolor=black,
}

\nocopyright 

\title{Semantic Complete Scene Forecasting from a 4D Dynamic Point Cloud Sequence}
\author{
    Zifan Wang\equalcontrib\textsuperscript{\rm 1,3},
    Zhuorui Ye\equalcontrib\textsuperscript{\rm 1,3},
    Haoran Wu\equalcontrib\textsuperscript{\rm 1,3},
    Junyu Chen\textsuperscript{\rm 1,3},
    Li Yi\textsuperscript{\rm 1,2,3}
}

\affiliations{
    \textsuperscript{\rm 1}Tsinghua University,~
    \textsuperscript{\rm 2}Shanghai Artificial Intelligence Laboratory,~
    \textsuperscript{\rm 3}Shanghai Qi Zhi Institute

    \{wzf22,yezr21,wuhr20,junyu-ch21\}@mails.tsinghua.edu.cn,~ ericyi@mail.tsinghua.edu.cn
}

\usepackage{bibentry}

\begin{document}



\twocolumn[{%
\renewcommand\twocolumn[1][]{#1}%
\maketitle

\vspace{-0.5cm}
\begin{figure}[H]
\hsize=\textwidth
  \centering
    \includegraphics[width=2\columnwidth]{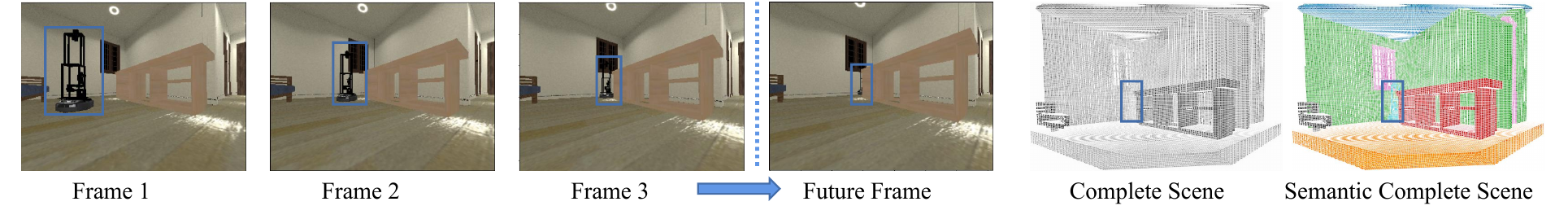}
  \caption{
\textbf{Semantic Complete Scene Forecasting.}
Combining past observations of an interaction scene, we can forecast the robot is about to walk behind a desk and will be occluded. Specifically, the SCSF task consumes a 4D point cloud sequence(here RGB is just to intuitively show the scene contents) and forecasts the complete scene with semantic labels in the whole space in the next frame. \revise{The SCSF task gives a comprehensive understanding of the future scene as shown in the last two figures.}}
\label{fig:intro}
\end{figure}
}]

\begin{abstract}
We study a new problem of semantic complete scene forecasting (SCSF) in this work. Given a 4D dynamic point cloud sequence, our goal is to forecast the complete scene corresponding to the future next frame along with its semantic labels. To tackle this challenging problem, we properly model the synergetic relationship between future forecasting and semantic scene completion through a novel network named SCSFNet. SCSFNet leverages a hybrid geometric representation for high-resolution complete scene forecasting. To leverage multi-frame observation as well as the understanding of scene dynamics to ease the completion task, SCSFNet introduces an attention-based skip connection scheme. To ease the need to model occlusion variations and to better focus on the occluded part, SCSFNet utilizes auxiliary visibility grids to guide the forecasting task. To evaluate the effectiveness of SCSFNet, we conduct experiments on various benchmarks including two large-scale indoor benchmarks we contributed and the outdoor SemanticKITTI benchmark. Extensive experiments show SCSFNet outperforms baseline methods on multiple metrics by a large margin, and also prove the synergy between future forecasting and semantic scene completion. The project page with code is available at 
\href{https://scsfnet.github.io}{\textcolor{magenta}{scsfnet.github.io}}.

\end{abstract}


\section{Introduction}
\label{sec:intro}

Visual forecasting has aroused a wide spectrum of interests in the computer vision community, especially for RGB videos\cite{zhang2019z, hu2019novel, luc2020transformation, gao2019disentangling, hu2020probabilistic}. Recently point cloud sequence forecasting~\cite{wen2022point, sun2020novel, deng2020temporal, weng2021inverting, mersch2022self, wang2022sequential} has also obtained many attention. Forecasting the future geometric observation can help an intelligent agent plan its behavior accordingly, which in turn greatly benefits a wide range of downstream applications in autonomous driving, robotics, and augmented reality.

Existing works on point cloud forecasting~\cite{wen2022point, sun2020novel, deng2020temporal, weng2021inverting, mersch2022self} mostly focus on forecasting raw future observation. The observation only provides a partial view of the dynamic scene, insufficient for a robot to perform low-level tasks like grasping and obstacle avoidance~\cite{song2017semantic} that require complete geometric understanding. Also, the raw future point cloud lacks semantic meaning, crucial for high-level tasks like object retrieval. Therefore, we propose a new task named \textbf{Semantic Complete Scene Forecasting (SCSF)} in this work. Given previous $N$ frames of an egocentric 4D dynamic point cloud sequence, the task is to predict the complete scene along with its semantic labels corresponding to the next future frame.

The SCSF problem is very challenging since it requires both future forecasting and semantic scene completion~\cite{song2017semantic}, or more intuitively, completing and segmenting something unknown. Previous works on point cloud forecasting~\cite{ilg2017flownet,dosovitskiy2015flownet} often formulate it as a flow prediction problem where future frames are treated as the deformation of earlier partial observations. However, they cannot provide a complete scene understanding which is crucial for occlusion inference or robot decision-making. On the other hand, works on semantic scene completion~\cite{roldao20223d} consider only static scenes. Previous research treated forecasting and completion as two separate problems and no works exploited their connections deeply.

Our key observation is that \textbf{future forecasting and semantic scene completion are synergetic to each other} and can lead to huge gains mutually if modeled properly. On one hand, future forecasting models the underlying scene dynamics and improves motion understanding. Such understanding allows aggregating multiple observations across time for a more complete geometric understanding. On the other hand, a complete scene understanding allows future forecasting to focus on the object movements without bothering with sampling noise or occlusion variations faced by traditional point cloud forecasting methods. Thus semantic scene completion can greatly improve future forecasting and provide a comprehensive understanding of the scene.


Based on the observation, we jointly solve the forecasting and the semantic completion in 4D. We struggle to answer three questions: How can we forecast the complete future scene with high resolution? How can we leverage the multi-frame partial observations to solve the completion task? How can we leverage the prior of empty and occluded space, which is usually ignored in point cloud forecasting problems, to tackle the forecasting task?
We address the above issues through a novel network named \textbf{SCSFNet}.

Specifically, to address the first issue, unlike traditional semantic scene completion methods that use low-resolution dense voxel representation, SCSFNet leverages a hybrid representation combining the benefits of sparse voxels, dense voxels, and implicit fields for high-resolution predictions. It employs a 4D sparse voxel encoder for fine-grained input information, uses dense voxels for flexible low-resolution structure generation, and predicts an implicit field from the low-resolution dense voxel grid for high-resolution predictions.
To address the second issue, we utilize 4D sparse convolutions to aggregate multi-frame observations in the past and introduce attention-based skip connections in SCSFNet. This innovative design improves information propagation and occlusion inference for more precise future scene forecasting. 

To address the third issue, we introduce an auxiliary task of forecasting a future visibility grid from past ones. This is a much easier task than predicting an entire high-resolution scene. The visibility grids' ambient information effectively aids in scene understanding.

To confirm the effectiveness of our method for intelligent agents, we construct two large-scale SCSF benchmarks,\textbf{ IGPLAY} focusing on robot interaction with indoor objects, and \textbf{IGNAV} focusing on robot social navigation with other dynamic agents in scenes. 
Besides the two indoor datasets, we also use SemanticKITTI \cite{behley2019iccv} to verify our method for outdoor scenarios. Extensive experiments demonstrate that by jointly modeling geometry forecasting and semantic completion, SCSFNet can outperform baseline methods by a large margin ($11.6\%$ CD relative improvements on the point cloud forecasting task and $10.8\%$ mIoU improvement on the SCSF task, on the IGPLAY dataset).

To summarize, our main contributions are fourfold: i) We propose a new task of semantic complete scene forecasting from a 4D dynamic point cloud sequence; ii) We present SCSFNet to exploit the synergy between forecasting and semantic scene completion to effectively solve such a new task; iii) To evaluate our method, we introduce IGPLAY and IGNAV, two large-scale 4D egocentric vision datasets with complete geometry and semantic annotations covering abundant indoor scenes and rich dynamics; \revise{iv) Our experiments verify the effectiveness of our method and prove the synergy between forecasting and semantic scene completion.}

\begin{figure*}[t]
  \centering
  \includegraphics[width=2\columnwidth]{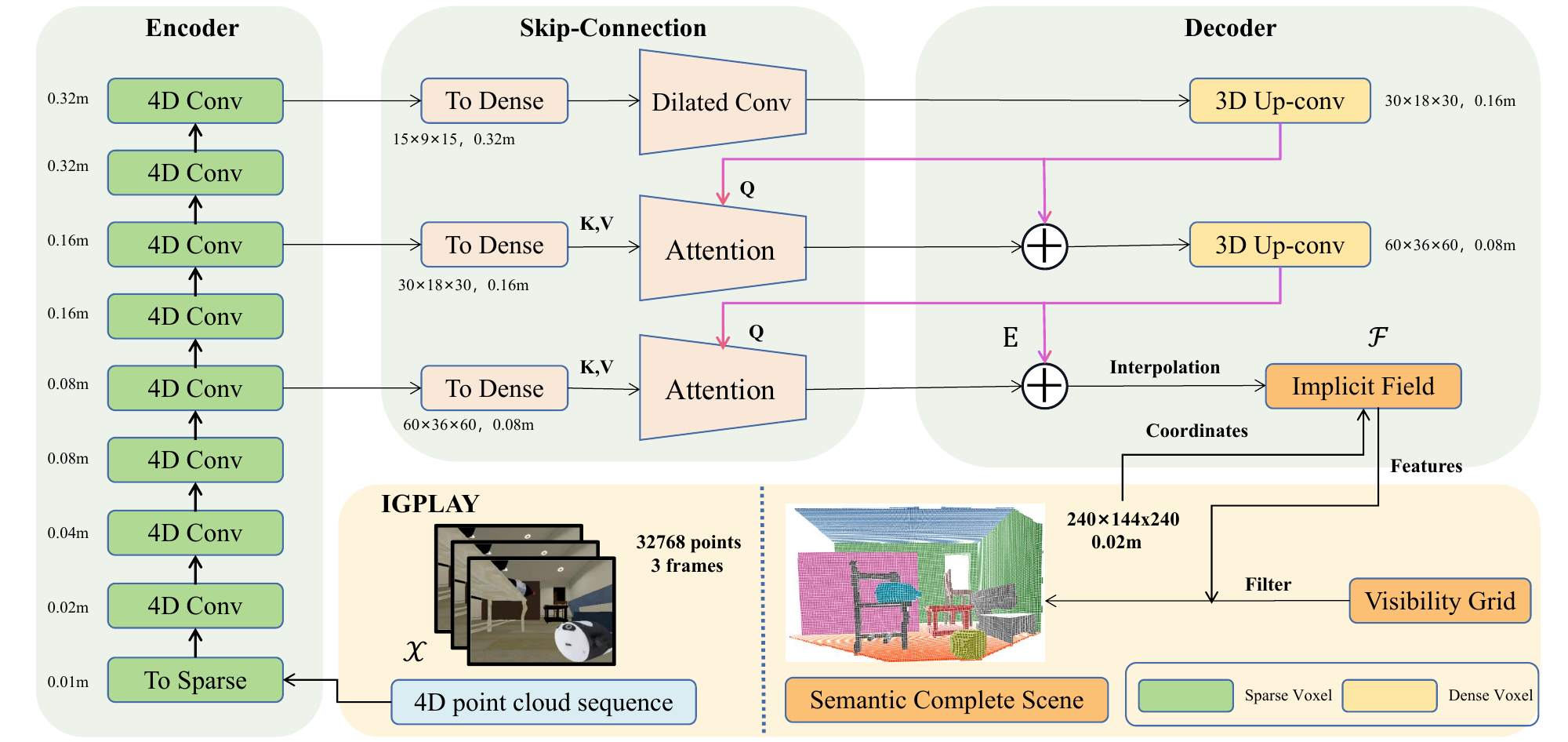}
  \caption{\textbf{The overview of SCSFNet.} First, a 4D point cloud sequence $\mathcal X$ will be transformed into 0.01m sparse voxels. The 4D sparse convolution encoder then generates spatial-temporal features at different scales (0.08m, 0.16m, and 0.32m). They are filled into 3D dense voxel grids and processed with attention-based skip connections at each scale. Finally, the model produces an implicit field $\mathcal F$ based on a low-resolution voxel grid $E$ (0.08m), which is then used to generate a high-resolution voxel output (0.02m) filtered by the visibility grid for evaluation.}
  \label{fig:overview}
  \vspace{-0.3cm}
\end{figure*}

\section{Related Work}
\label{sec:related_work}
{\bf 4D Sequential Point cloud Forecasting.} 
Point cloud forecasting is crucial for understanding scene geometry and motion dynamics\cite{fan2021point,wen2022point}.
Various methods have been employed in different scenarios. 
TLFPAD \cite{deng2020temporal} proposed using scene stream embedding to model past point cloud frames' temporal relationships for future frame prediction.
Mersch \textit{et al.}\shortcite{mersch2022self} proposed to use 3D convolution to jointly learn the spatial-temporal features of the input point cloud sequence.
These approaches primarily emphasize visible surfaces rather than the entire scene.
Occlusion4d \cite{vanhoorick2022revealing} introduced a framework to estimate 4D visual representations from monocular RGB-D video, which encodes point clouds into a continuous high-resolution representation.
Unlike Khurana \textit{et al.}\shortcite{khurana2023point}, whose work attained SOTA point cloud forecasting using sensor extrinsics, our approach doesn't rely on camera poses or perfect odometry for scene alignment.
Instead, we aim to jointly solve forecasting and semantic completion in the egocentric view across both indoor and outdoor environments.

{\bf Semantic Scene Completion.} 
Semantic Scene Completion \cite{song2017semantic} has gained significant momentum in the research community due to the unresolved challenges it faces \cite{li2019rgbd, wang2019forknet, liu20203d, chen20203d, cheng2021s3cnet, rist2021semantic}. \textbf{This task's output is a 3D voxel grid with a semantic label for each voxel.} SISNet \cite{cai2021semantic} aims to deduce detailed shape information and nearby objects of similar categories. Without extra instance labels, SCSFNet can also obtain detailed geometric data using the implicit field based on the coarse voxel grid.
Most works in this field only aim to process 3D static scenes like NYU and SUNCG \cite{silberman2012indoor, song2017semantic, firman2016structured, xie2020grnet, Matterport3D} without any temporal information. We contribute two 4D dynamic indoor datasets IGPLAY and IGNAV.
As for outdoor scenarios, SemanticKITTI \cite{behley2019iccv} is a dataset suitable for Point Cloud Forecasting and 3D Semantic Scene Completion, allowing direct use of the former's input and the latter's ground truth for SCSF.


\indent {\bf Hybrid Geometric Representations.} Hybrid geometric representations \cite{ali2018hybrid,zhang2020h3dnet,song2020hybridpose} has aroused great interest  \cite{dourado2022data,xu2019disn,peng2020convolutional} for various downstream tasks. 
Convolutional Occupancy Networks \cite{peng2020convolutional} combines convolutional encoders with implicit occupancy decoders for 3D reconstruction.
Our SCSFNet further utilizes implicit fields for 4D complete forecasting.
GRNet \cite{xie2020grnet} and SCSFNet both leverage point cloud input and intermediate 3D voxel grids, but SCSFNet  decodes infinite implicit fields unlike GRNet converting the voxels back to point clouds explicitly with a limited resolution. While Liu \textit{et al.}\shortcite{liu2020neural} leverages self-pruning and sparse voxels while overfitting a scene, we instead use dense voxels enriched with skipped point cloud features to dynamically determine scene geometry.


\section{Method}

\addition{We first define the \textbf{Semantic Complete Scene Forecasting (SCSF)} task formally.
Let $P_t \in \mathbb{R}^{N \times 3}$ and $F_t \in \mathbb{R}^{N \times C}$ denote the coordinates and features of the $t$-th frame in a point cloud sequence, where $N$ and $C$ denote the number of points and feature channels. 
Given a point cloud sequence $\mathcal X=([P_1,F_1],\dots, [P_L,F_L])$ as input, the SCSFNet initially produces a low-resolution voxel feature grid $E\in \mathbb R^{X\times Y\times Z\times D}$, where $D$ represents the feature dimensions, and $X,Y,Z$ denote the dimensions of the low-level grid. 
Subsequently, Based on $E$, the SCSFNet then conducts interpolations and yields an implicit field $\mathcal F: \mathbb R^3\to \mathbb Z$, 
describing the semantic label at an arbitrary point in the next frame. To generate a specific high-resolution voxel grid, we can obtain semantic occupancy of each voxel by querying their central coordinates in $\mathcal F$.}



Our method, SCSFNet, uses an egocentric 4D point cloud sequence with $N$ frames to predict future scenes with semantic information. It employs an encoder-decoder structure and hybrid geometric representations for high-resolution forecasting. Like U-Net~\cite{ronneberger2015u}, SCSFNet uses skip connections between the encoder and decoder to retain high-resolution input information. However, challenges arise in its design. Traditional scene completion methods use dense voxel grids in the decoder, resulting in a resolution limit. It is also complex to allow skip connections between a partial input and a complete prediction at different resolutions. Further, previous point cloud forecasting methods solely operate on a point cloud representation, making it difficult to utilize ambient information such as empty or occlusion space distribution.

To address the three challenges mentioned above, we utilize a hybrid geometric representation for high-resolution generation, design attention-based skip connections for features from partial to complete, and also introduce visibility grids to leverage ambient information. In the following, we will explain how we leverage a hybrid geometric representation to design SCSFNet in Section~\ref{subsec:hybrid}. Then we introduce our attention-based skip connections and visibility grids in Section~\ref{subsec:skip} and Section~\ref{subsec:visibility} respectively. 

\subsection{SCSFNet with a Hybrid Geometric Representation}
\label{subsec:hybrid}
Semantic complete scene forecasting involves predicting future geometry and completing the scene. Traditional point cloud forecasting methods focus on point cloud representation, making it difficult to recover complete scene geometry. In semantic scene completion, the dense voxel grid is commonly used due to its ease in generating new structures, but its resolution constraint limits its use for tasks requiring detailed geometric understanding in large scenes. Hence, we utilize a hybrid geometric representation that combines the advantages of point clouds, sparse and dense voxels, and implicit fields for high-resolution predictions.

We design SCSFNet, an encoder-decoder structure that encodes the high-resolution point cloud sequence input and decodes an implicit field with infinite resolution based on a low-resolution voxel grid, as shown in Figure~\ref{fig:overview}.

SCSFNet effectively processes input with minimal resolution loss by filling points into small voxels and utilizing Minkowski Engine~\cite{minkovskiengine} for 4D sparse convolutions.
By incrementally increasing the convolution stride, we enlarge voxel sizes and obtain multi-scale features. Compared with 4D point cloud encoders like P4Transformer~\cite{fan2021point}, the sparse voxel-based encoder aligns better with a dense voxel decoder, which is commonly used for semantic scene completion and structure generation.



After our 4D sparse convolution encoder extracts spatial-temporal features at different scales, we fill the aggregated features of sparse voxels into a 3D dense voxel grid with the same resolution at each scale. This information is fed to the decoder through skip connections, similar to U-Net~\cite{ronneberger2015u}.
More details about our attention-based skip-connections are in Section~\ref{subsec:skip}.

The dense voxel exploited by our decoder makes it very easy to generate new geometry through up-convolutions. We leverage several 3D up-convolution layers to lift the resolution of the voxel grid to a higher scale. However, due to the restriction of memory, we cannot reach high resolution with such a dense voxel representation. Inspired by Convolutional Occupancy Networks~\cite{peng2020convolutional}, we further convert such a dense voxel feature grid into a semantic occupancy field to enable infinite resolution in the output. In particular, for an arbitrary point $p\in\mathbb{R}^3$, we obtain its feature by conducting a trilinear interpolation using the features of nearby voxels. Then the point feature goes through a Multilayer Perceptron~(MLP) to reach a semantic occupancy prediction where the label is either empty or a semantic class.

Combining a sparse voxel encoder, a dense voxel decoder, and an implicit field, we can encode the high-resolution input with little information loss while at the same time forecasting a complete scene with infinite resolution.

During the training phase, we use a loss based on the high-resolution voxel ground truth. This ground truth has a higher resolution compared with the dense voxel grid on which the implicit field is based. We randomly sample $10^5$ points from all central coordinates of ground truth voxels on the fly during training and evaluate the semantic occupancy values of these points.
We use a cross-entropy loss $\mathcal L_{\text{high}}$  to supervise.
During inference, we can forecast a high-resolution dense voxel grid by simply querying the semantic occupancy of the central coordinate of each voxel. \addition{It is noted that the SCSFNet is trained and evaluated using points from the ground truth voxel centroids, rather than
exact implicit functions. This approach is chosen due to the absence of ground truth meshes in existing datasets with high-resolution information, rendering it unfeasible to acquire precise implicit functions for both training and evaluation purposes. Moreover, this methodology aligns with common practices in baseline methods, which often perform experiments using point clouds, ensuring a fair and meaningful comparison.}

\begin{figure}[t]
  \centering
  \includegraphics[width=\columnwidth]{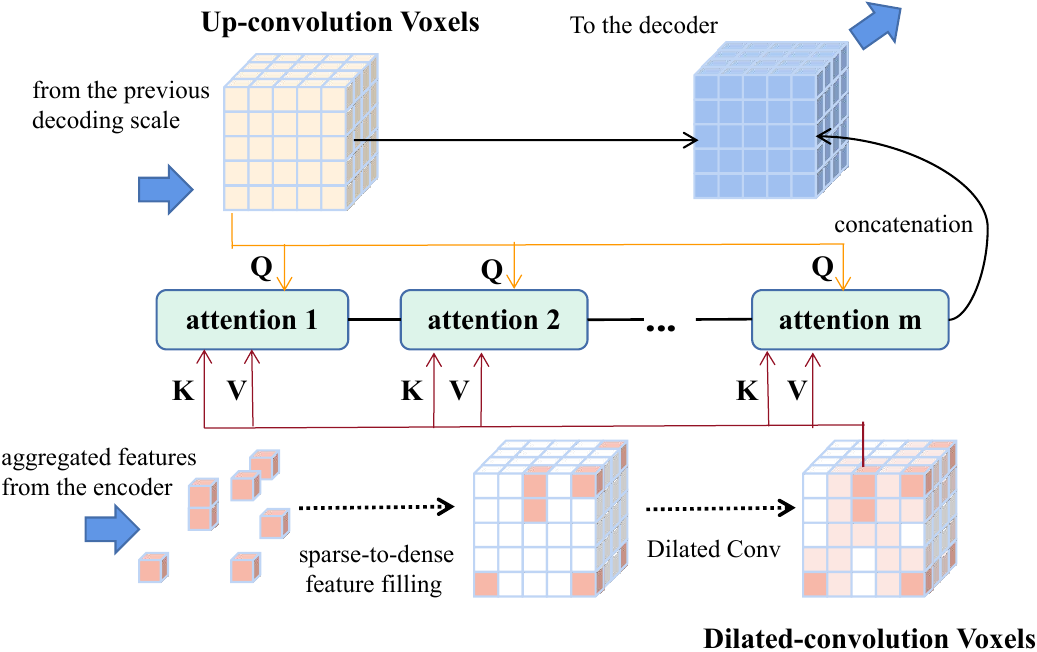}
  \vspace{-0.3cm}
  \caption{\textbf{Illustration of attention-based skip connections.} Dilated-convolution voxels from the encoder form the key and value. Up-convolution voxels from the previous scale form the query. Finally, we concatenate voxel features from multi-head cross attention and feed them to the decoder.}
\label{fig:skip}
\vspace{-0.6cm}
\end{figure}

\subsection{Attention-Based Skip Connection}
\label{subsec:skip}
The naive encoder-decoder structure struggles to transfer detailed information in the bottleneck layer, making dense prediction tasks tough. Researchers have suggested using skip connections in a U-Net structure~\cite{ronneberger2015u} to shortcut high-resolution features, which has improved tasks like semantic segmentation. However, the modality gap between the point cloud input and the dense voxel grid output presents a more significant challenge. This can be partially addressed by voxelizing the input and using a sparse voxel encoder, but transitioning from partial observations to complete predictions remains difficult.

To address the above challenge, we introduce the attention-based skip connections as shown in Figure~\ref{fig:skip}. Given a set of sparse voxels at a certain scale of our encoder and a dense voxel grid at the corresponding scale of the decoder, our goal is to feed the sparse voxel features to the dense voxels in a similar fashion to traditional skip connections. An intuitive approach would be to simply add sparse voxels from the encoder to the dense voxels of the decoder. However, this method provides no information to the majority of the voxels, which are crucial for accurately forecasting a complete scene. Our attention-based skip connections are designed to alleviate this problem. 


Specifically, we obtain 3D sparse aggregated features in the encoder, which has consolidated spatial-temporal details at different scales. These features are then filled in dense voxel grids, but only a limited subset of voxels contain information. To diffuse the encoded details efficiently, we utilize dilated convolutions. The outcomes of this step are referred to as \textbf{``dilated-convolution voxels''} as shown in Figure~\ref{fig:skip}. And we call the voxels produced through 3D up-convolutions in the decoder as \textbf{``up-convolution voxels''}.

At the most coarse-grained scale, dilated-convolution voxels are directly processed with up-convolutions to enter the next scale. 
At other scales, we combine dilated-convolution voxel features with positional embeddings, generating key-value pairs ${(K_j, V_j)}$ for key and value matrices $K$ and $V$ respectively. Positional embeddings are computed using the Fourier basis of voxel centroid positions through an MLP. For the $i$-th up-convolution voxel at the previous scale, a query $Q_i$ is derived from its voxel feature and positional embedding. This allows us to compute the skipped feature for the $i$-th up-convolution voxel using $\text{softmax}(\frac{Q_iK^T}{\sqrt{d_k}})V$, where $d_k$ is the dimension of $Q_i$. In practice, we utilize multi-head cross-attentions and concatenate them to the final skipped features.
Following U-Net's common practice, we combine skipped features with original up-convolution voxel features, which are then forwarded to the decoder.

Such design allows feeding high-resolution features directly from the encoder to every voxel in the decoder, facilitating implicit geometry completion and scene forecasting.

\subsection{Visibility Grid}
\label{subsec:visibility}
\begin{figure}[t]
  \centering
  \includegraphics[width=1\columnwidth]{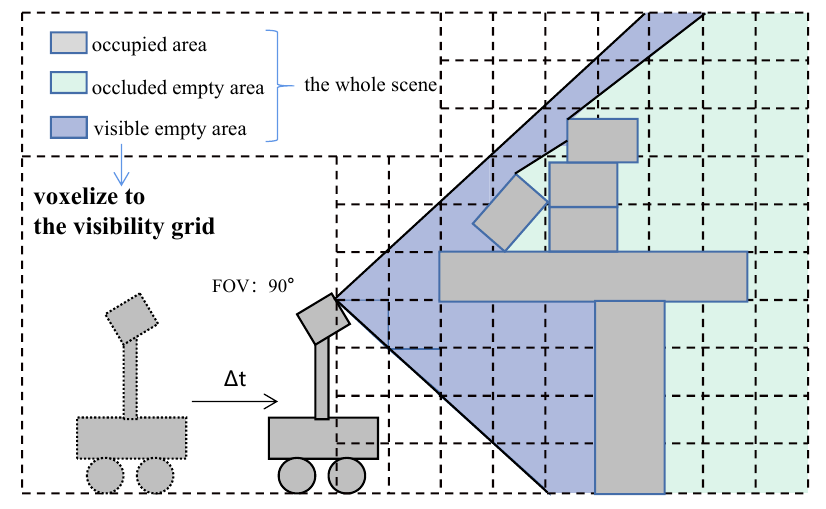}
  \vspace{-0.3cm}
  \caption{\textbf{Illustration of the visibility grid.} An iGibson robot is nearing a table. Forecasting visible empty areas is much easier than the whole empty areas. We specifically forecast the future visibility grid to assist the SCSF task in predicting the entire scene.}
\label{fig:vg}
\vspace{-0.6cm}
\end{figure}

There is a strong ambient space prior in a point cloud from which we can extract the visible empty area: if we shoot a ray from the depth camera to a depth point in the scene, the space between the camera and the point should be empty; when we further extend this ray, it would reach the occluded space. This information is critical for a complete scene forecasting task since our main job is to inpaint the occupied area but not the visible empty space, as shown in Figure~\ref{fig:vg}. This information is rarely used in traditional point cloud forecasting tasks since with a point cloud representation, such ambient information is hard to encode. Thanks to the hybrid representation, we can emphasize such important information for the SCSF task.

In particular, we introduce a dense visibility grid to model the information. By voxelizing each frame of the input point cloud sequence and projecting rays to identify visible voxels, we extract valuable visibility data. The size of the visibility grid matches the high-resolution ground truth voxel dimensions. This visibility grid is intended to complement our predicted results, but the challenge is that we know nothing about the scene in the next frame.

To address this issue, we introduce an auxiliary task of forecasting a future visibility voxel grid from past visibility grids. This task is much easier than recovering a complete scene with semantics in infinite resolution, so we can train a relegated version of SCSFNet to obtain a high-resolution visibility voxel grid with binary labels.
Then we use the grid as an additional mask to filter out visible empty areas for the main SCSFNet output, which can ease the need to model occlusion variations when solving the challenging SCSF task.



\section{Datasets}
In order to train and evaluate our model, and to demonstrate the ability of our model for 4D completion and forecasting, we require sequential point clouds with both 3D semantic scene completion annotations and underlying dynamics that can be used for forecasting. For this purpose, we contribute two high-quality synthetic datasets IGPLAY and IGNAV using the Interactive
Gibson Simulator (iGibson)~\cite{igibson} that runs on top of the
pyBullet~\cite{pybullet}. To generate the simulated scenes suitable for our task, 
we specify the desired configuration of the environment in the logic language BDDL. With the BDDL description and a list of scene names, the logic states implemented in iGibson 2.0 provide a mechanism that facilitates the generation of simulated scenes with various objects. 
In our datasets, we provide 2D pictures (RGB, normal and semantics), 3D point clouds, 3D meshes, 3D visibility grids 
and 3D ground truth voxel grids. The train/test split is $80\%/20\%$. 
The time unit in iGibson is ``timestep" instead of ``second", so we just specify how far the robot and objects move in each timestep by iGibson interfaces.
We provide more statistics in the supplementary.
Besides the two indoor datasets we provide, the outdoor SemanticKITTI dataset is also suitable for our task. 
\addition{Given the limited options
for real-world datasets meeting these criteria, we find SemanticKITTI to be the only suitable choice, which consists
of large-scale 4D point cloud sequences with completion
and semantic annotations. We hope to extend the evaluation
of our approach to suitable real-world datasets released in
the future.}

\subsection{IGPLAY \& IGNAV}
We present the IGPLAY dataset by capturing 1,000 scenes
lasting 10 timesteps each, where 
a viewer interacts with various toys among the furniture.
Each scene in IGPLAY contains 10 semantic classes in a great variety.
When the viewer (an iGibson robot) moves and interacts with objects, partial and complete occlusions happen between the furniture and objects.
This dataset supports effective learning
for  4D dynamic scene completion and 
 forecasting.
 
While IGPLAY already exhibits many features suitable for 4D forecasting, the whole dynamics of the scene only come from the viewer's movement and interaction with the objects. To provide a new perspective of the underlying dynamics, we provide a new dataset called IGNAV with several active robots navigating in the scene. 
Each scene in IGNAV contains 9 semantic classes. We captured 600 scenes lasting 10 timesteps each, where several robots move around the scene in random speeds, with another robot watching them.

\subsection{SemanticKITTI}
SemanticKITTI \cite{behley2019iccv} is a very challenging and well-known large-scale outdoor dataset collected by autonomous cars. Various occlusions by trees or other cars make the dataset suitable for evaluating our SCSF task.

\begin{table*}[t]
\setlength{\tabcolsep}{6mm}
\centering\small
\resizebox{\textwidth}{!}{
\begin{tabular}{|l| c c | c c | c c |} 
\hline
 &  \multicolumn{2}{c|}{IGPLAY} & \multicolumn{2}{c|}{IGNAV} & \multicolumn{2}{c|}{SemanticKITTI} \\ \hline
Method   & IoU &   mIoU  & IoU &  mIoU & IoU & mIoU \\ 
\hline
Occlusion4d~\cite{vanhoorick2022revealing}   (CSF) & 39.7 & - & 45.0 & -  & 16.4 & - \\ 

Occlusion4d~\cite{vanhoorick2022revealing} (SCSF) & 40.8  & 23.8 & 46.1 & 24.6 & 17.1 & 6.4  \\

TLFPAD~\cite{tlfpad} (CSF) & 48.2  & - & 60.2 & - & 32.3 & -   \\

TLFPAD~\cite{tlfpad} (SCSF) & 48.6  & 25.8 & 60.7 & 25.9 & 32.6 & 9.5 \\

ST3DCNN~\cite{mersch2022self} (CSF) & 49.1  & - & 61.4 & - & 28.8 & -   \\

ST3DCNN~\cite{mersch2022self} (SCSF) & 49.4  & 25.5 & 62.9 & 26.1 & 29.5 & 9.6 \\

\hline

SCSFNet (CSF)  & 53.9  & - & 63.6 & - & 33.7 & - \\

SCSFNet (SCSF)  & \textbf{56.5}  & \textbf{36.3} & \textbf{69.5} & \textbf{39.9}  & \textbf{34.5} & \textbf{16.1} \\

\hline
\end{tabular}
}
\vspace{-0.2cm}
\caption{\textbf{SCSF Results on IGNAV, IGPLAY, and SemanticKITTI}. We compare SCSFNet with baselines on both the CSF and the SCSF task using IoU and mIoU (in percentages). Our method significantly outperforms baselines, especially in mIoU.}

\label{tab:scsf-task_iccv}
\vspace{-0.2cm}
\end{table*}

\begin{table*}[t]
\setlength{\tabcolsep}{6mm}
\centering\small
\resizebox{\textwidth}{!}{
\begin{tabular}{|l| c c | c c | c c |} 
\hline
 &  \multicolumn{2}{c|}{IGPLAY} & \multicolumn{2}{c|}{IGNAV} & \multicolumn{2}{c|}{SemanticKITTI} \\ \hline
Method   &  IoU(M) &   IoU(S)  & IoU(M) &  IoU(S) & IoU(M) & IoU(S) \\ 
\hline

Occlusion4d~\cite{vanhoorick2022revealing} (SCSF) &  6.3 & 28.2 & 4.4 & 27.1 & 3.4 & 7.8  \\

TLFPAD~\cite{tlfpad} (SCSF) & 11.9  & 29.3 & 2.1 & 28.8 & 4.9 & 11.7 \\

ST3DCNN~\cite{mersch2022self} (SCSF)  & 9.3  & 29.6 & 0.0 & 29.3 & 4.8 & 11.9 \\

\hline

SCSFNet  (SCSF) & \textbf{28.2}  & \textbf{38.4} & \textbf{18.3} & \textbf{42.6}  & \textbf{12.4} & \textbf{17.8} \\

\hline
\end{tabular}
}
\vspace{-0.2cm}
\caption{\textbf{SCSF Results on dynamics of IGNAV, IGPLAY, and SemanticKITTI}. In the same setting as Table \ref{tab:scsf-task_iccv}, we compare Movable IoU and Static IoU (in percentages) for the SCSF task. Our method outperforms baselines, especially in Movable IoU.}

\label{tab:scsf-task_iccv2}
\vspace{-0.2cm}
\end{table*}

\section{Experiments}
We evaluate our proposed methods on two indoor synthetic datasets (IGPLAY and IGNAV) and one outdoor real-world dataset (SemanticKITTI).

For IGPLAY and IGNAV, the dimensions of the 3D space are 4.8m horizontally, 2.88m vertically, and 4.8m in depth. We use 3 RGB frames (interval: 1 iGibison timestep) as the input and a $240 \times 144 \times 240$ volume with grid size 0.02m as the ground truth, which is similar to \cite{song2017semantic}.

For SemanticKITTI, the dimensions of the 3D space are 51.2m ahead of the car, 25.6m to every side of the vehicle, and 6.4m in height. We input three frames of point clouds (0.2s interval) and use a $256 \times 256 \times 32$ volume with a 0.2m grid size as the ground truth, provided by SemanticKITTI for the semantic scene completion benchmark.

The three datasets have distinct differences. IGPLAY and IGNAV use a $480 \times 640$ image from a depth camera, convertible to a point cloud, while SemanticKITTI uses a raw LiDAR point cloud with varying points. IGPLAY and IGNAV contain more dynamic objects, whereas SemanticKITTI features larger spaces and complex real-world details. Well managing these datasets shows the versatility of SCSFNet.

\subsection{Evaluation Metrics}
To evaluate forecasting results, we obtain high-resolution voxels or point clouds from implicit fields. We evaluate scene completion quality using voxel-level intersection over union (\textbf{IoU}) between predicted and ground truth labels. If labels contain semantics, we assess semantic scene completion quality using mean IoU across all classes (\textbf{mIoU}). For point clouds, we use point-level Chamfer Distance (\textbf{CD}).

\subsection{Evaluating SCSFNet on the CSF Task and the SCSF Task}

\begin{figure*}[t]
  \centering
  \includegraphics[width=2\columnwidth]{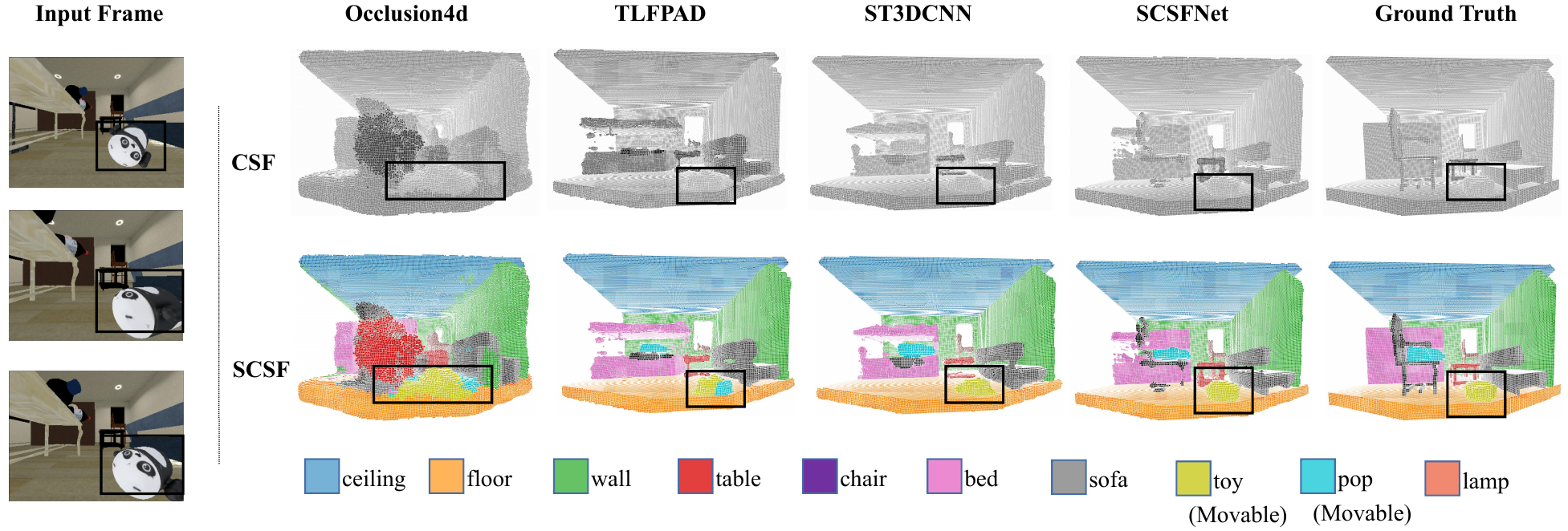}
  \caption{\textbf{Visualization results in IGPLAY.} In this sequence, an iGibson robot is approaching a panda toy on the ground and is going to hit it. Occlusion4d predicts the fuzzy shape of the toy. TLFPAD predicts half of the toy as the pop class. In comparison, SCSFNet can get relatively accurate geometry and semantics of the toy compared with the ground truth.}
  \label{fig:igplay_vis}
\vspace{0.3cm}
  \includegraphics[width=2\columnwidth]{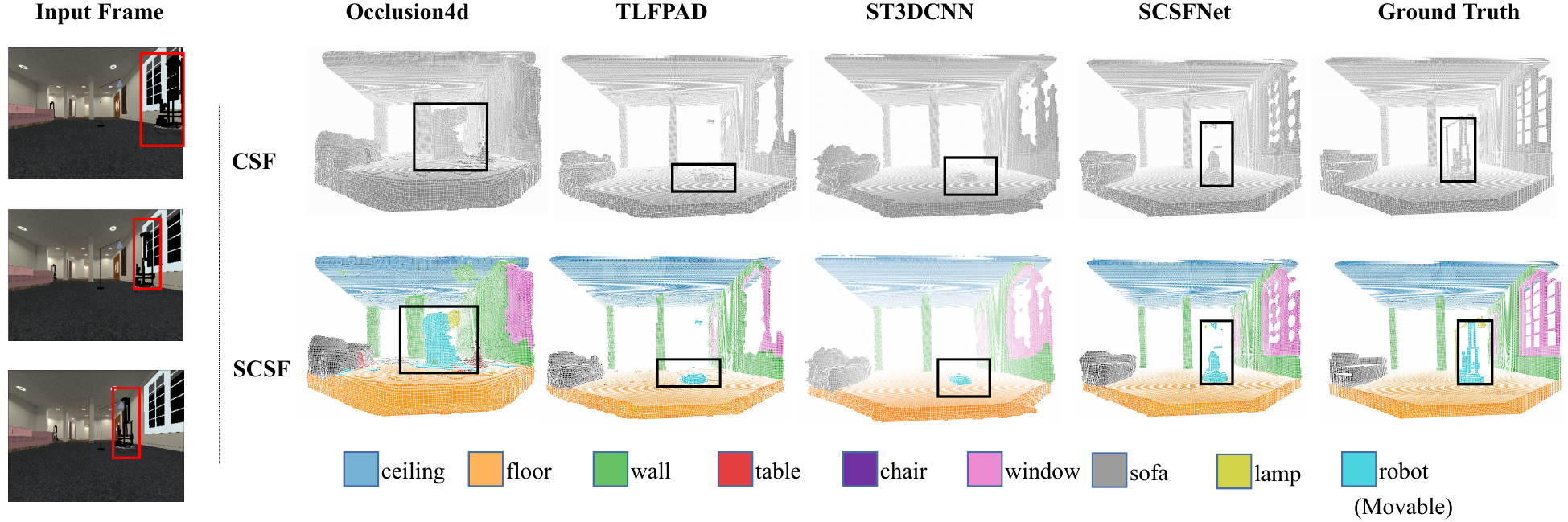}
  \caption{\textbf{Visualization results in IGNAV.} In this sequence, an iGibson robot is watching another robot beside the window going left. SCSFNet can gain more structures of this ``slim'' robot and other objects such as the window and the sofa.}
  \label{fig:ignav_vis}
\end{figure*}


For the SCSF task defined above, we need to forecast the semantic scene completion. We can also delegate SCSF into the Complete Scene Forecasting (CSF) task, where we only forecast the scene completion as a binary classification.

\textbf{Baselines.} Since this paper is pioneering the SCSF task, there are no existing methods for comparison. We devise baselines from existing modern methods. The input is a 3-frame point cloud sequence and the output is the future complete scene.(1) \textit{Occlusion4d} \cite{vanhoorick2022revealing} is designed for estimating 4D visual representations and occlusions from RGB-D video. It can be trained end-to-end for the SCSF task. (2)\textit{TLFPAD}~\cite{tlfpad}, a FlowNet3D-based point cloud forecasting method is used for partial forecasting. Therefore we first forecast a partial point cloud of the next frame using TLFPAD, and then train a SCSFNet exclusively for (semantic) scene completion on the forecasted point cloud. (3) \textit{ST3DCNN}~\cite{mersch2022self}, a CNN-based point cloud forecasting method achieves SOTA (without sensor extrinsics) on the KITTI-Odometry dataset~\cite{Geiger2012CVPR}. We follow the same two-stage training approach as TLFPAD.

\textbf{Comparison results between SCSFNet and baselines.} SCSFNet outperforms baselines on three different datasets by a large margin as shown in Table \ref{tab:scsf-task_iccv}. 
Thanks to the hybrid geometric representation and the implicit field, SCSFNet obtains high-resolution forecasting on both indoor and outdoor scenes. We observe Occlusion4d perform not well for whole-scene forecasting. Compared with SCSFNet on IGPLAY~(SemanticKITTI), ST3DCNN has a $7.1\%(5\%)$ reduction in IoU but a $10.8\%(6.5\%)$ reduction in mIoU. Separating forecasting and semantic completion into two stages hampers completion and extraction of semantics, and since semantics prediction is harder and requires more information, the drop in mIoU is a bit larger. SCSFNet excels by jointly incorporating semantic scene completion and forecasting. The results are visualized in Figure \ref{fig:igplay_vis}. 

\textbf{Comparison results between the CSF and SCSF Task.}
Two Adjacent rows in Table \ref{tab:scsf-task_iccv} compare the performance of the same method across the CSF task and SCSF task. Notably, SCSFNet displays improvements: a $2.6\%$ enhancement in IGPLAY, a $5.9\%$ enhancement in IGNAV, and a $0.8\%$ enhancement in SemanticKITTI. Baselines similarly exhibit improved IoU scores. This underscores the value of object semantic understanding for scene completion.

\textbf{Comparison results between movable and static classes.}
We also report IoU when evaluating movable or static classes in Table \ref{tab:scsf-task_iccv2}. A class is labeled as movable if the corresponding objects can move, such as robots in IGNAV or cars in SemanticKITTI.
This distinction allows us to emphasize a method's ability to comprehend scene dynamics and forecast movable classes.
 In the SCSF task, we categorize predicted voxels as either movable or static based on their predicted semantic labels. This allows us to calculate IoU for each category and subsequently determine the Average Movable IoU and Average Static IoU.
Compared to ST3DCNN on the SCSF task, our SCSFNet leads to $18.9\% (8.8\%)$, $18.3\% (13.3\%)$, $7.6\% (5.9\%)$ improvements in movable (static) IoU for IGPLAY, IGNAV, and SemanticKITTI respectively. 
Therefore, SCSFNet's concurrent forecasting and completion harness the advantageous interaction. This integrated approach improves the handling of dynamic objects, a critical factor in accurate future forecasting.
More details are provided in the supplementary.


\newcommand{\tablestyle}[2]{\setlength{\tabcolsep}{#1}\renewcommand{\arraystretch}{#2}\centering\footnotesize}
\newcolumntype{Y}{>{\centering\arraybackslash}X}

\begin{table}[h]
\setlength{\tabcolsep}{2mm}
\centering
\footnotesize
\tablestyle{5pt}{1.2}
\begin{tabularx}{1.0\linewidth}{lYY}
    \toprule
    \multicolumn{1}{c}{Methods} & \multicolumn{1}{c}{IGPLAY} & \multicolumn{1}{c}{IGNAV} \\
      
    \midrule
    TLFPAD \cite{tlfpad} & 0.045 & 0.043  \\ %
    ST3DCNN \cite{mersch2022self} & 0.043 & 0.037  \\ %
    \hline
    SCSFNet~(SCSF) & \textbf{0.038} & \textbf{0.030}  \\
    \bottomrule
\end{tabularx}
\caption{\textbf{Results on the point cloud forecasting of the next frame}. We report the Chamfer Distance of the point cloud of the next frame (lower is better, the unit is square meter). }
\vspace{-0.5cm}
\label{tables:partial}
\end{table}

\setlength{\tabcolsep}{2pt} 

\subsection{Completion Helps Point Cloud Forecasting}
\label{subsec:complete_to_forecast}
\revise{We evaluate SCSFNet on point cloud forecasting task to verify semantic scene completion helps forecasting.} 
Using iGibson's intrinsic and extrinsic parameters,  we project forecasted complete high-resolution voxels onto partial ones in IGPLAY and IGNAV, from which we can extract a fine-grained point cloud.

\textbf{\revise{Configurations.}} TLFPAD \cite{tlfpad} and ST3DCNN \cite{mersch2022self} are tailored to predict the point cloud of the future frame. We naturally train these two baselines end-to-end, using the next frame's point cloud as the ground truth. \revise{In comparison, SCSFNet is trained as usual (on SCSF task)}, and we use the above complete-to-partial projection to obtain a partial point cloud from the predicted complete scene. 

\textbf{Results.} By jointly considering semantic completion and forecasting, our SCSFNet outperforms ST3DCNN, improving CD from $0.043$ to $0.038$ in IGPLAY and from $0.037$ to $0.030$ in IGNAV, as shown in Table \ref{tables:partial}. 

\subsection{\revise{Forecasting Helps Semantic Completion}}
\label{subsec:forecast_to_complete}
\revise{We show that future forecasting benefits semantic scene completion in the Semantic Scene Completion task~\cite{song2017semantic}.} The Semantic Scene Completion task does not introduce any forecasting knowledge as described in Section \ref{sec:related_work}.

\newcolumntype{Y}{>{\centering\arraybackslash}X}

\begin{table}[t]
\setlength{\tabcolsep}{2mm}
\centering
\footnotesize
\tablestyle{5pt}{1.2}
\begin{tabularx}{1.0\linewidth}{lYY}
    \toprule
    \multicolumn{1}{c}{Methods} & \multicolumn{1}{c}{IoU} & \multicolumn{1}{c}{mIoU} \\
      
    \midrule
    SSC (from scratch) & 60.1 & 39.5  \\
    SSC (finetuned by SCSF) &\textbf{62.3} & \textbf{40.5} \\
    \bottomrule
\end{tabularx}
\caption{\textbf{Results on the Semantic Scene Completion task}. The two experiments are on IGPLAY. The first model is trained in the SSC setting from scratch and the second model is pre-trained in the SCSF setting and fine-tuned in the SSC setting.}
\vspace{-0.2cm}
\label{tables:forecast_completion}
\end{table}

\setlength{\tabcolsep}{2pt} 
\textbf{Configurations.}
We conduct two IGPLAY experiments. (1)Train a SCSFNet from scratch with 3 past frames to predict the complete scene in the last frame. (2) Train a SCSFNet for the SCSF task (using 3 past frames to predict the scene in the future frame), and finetune it to predict the complete scene in the last given frame.

\textbf{Results.} The SCSFNet finetuned from the SCSF task improves IoU by $2.1\%$ and mIoU by $1.0\%$, as shown in Table \ref{tables:forecast_completion}. This demonstrates the critical role of additional forecasting knowledge in semantic scene completion. The model, trained within the context of the SCSF task, acquires the ability to forecast, thereby offering deeper insights into scene completion.

\textbf{We establish the synergy between completion and forecasting by proving both directions in Section \ref{subsec:complete_to_forecast} and Section \ref{subsec:forecast_to_complete}.}

\subsection{Ablations}
Table \ref{tables:ablations} shows how different architectural choices affect our network's performance for the SCSF task in IGPLAY.

\textbf{Implicit representations help high-resolution forecasting.} An examination of our model without implicit representations involves using a low-resolution voxel-wise cross-entropy loss for supervision and excluding the implicit representation. When assessing this model against high-resolution ground truths, we up-sample the predicted low-resolution voxel grid to a high-resolution one through a simple copying mechanism. However, this up-sampling method results in a loss of the ability to capture fine-grained geometry, leading to a significant decrease of 9.5\% in Intersection over Union (IoU) and 9.9\% in mean Intersection over Union (mIoU). This underscores the crucial role of implicit representations as key components in our SCSFNet.

\textbf{Attention-based skip connections benefit spatial-temporal features. }When simply incorporating point filling into voxels as skip connections, as opposed to utilizing our well-designed cross-attention modules, SCSFNet exhibits a lower IoU and mIoU by 2.2\% and 4.8\%, respectively. This observation demonstrates that our attention-based skip connections play a pivotal role in enhancing spatial-temporal feature learning.

\newcolumntype{Y}{>{\centering\arraybackslash}X}

\begin{table}[t]
\setlength{\tabcolsep}{3mm}
\centering
\footnotesize
\tablestyle{2.5pt}{1.2}
\begin{tabularx}{1.0\linewidth}{lYY}
    \toprule
    \multicolumn{1}{c}{Methods} & \multicolumn{1}{c}{IoU} & \multicolumn{1}{c}{mIoU} \\
    \midrule
    No implicit fields & 47.0 & 26.4  \\ %
    No  attention-based skip connections & 54.3 & 31.5  \\
    No visibility grids & 55.7 & 34.3  \\
    P4T encoder \cite{fan2021point} & 50.9 & 29.5  \\
    Ours & \textbf{56.5} & \textbf{36.3}  \\ %
    \bottomrule
\end{tabularx}
\caption{\textbf{Ablation study of SCSFNet on IGPLAY for the SCSF task.} P4T means P4transformer.}
\vspace{-0.2cm}
\label{tables:ablations}
\end{table}

\setlength{\tabcolsep}{2pt} 

\textbf{Pretrained visibility grid is necessary for scene understanding.} This auxiliary task for the ambient space is easier to train than the main SCSF task. We can train a relegated version of SCSFNet to obtain a high-resolution
visibility voxel grid with binary labels and can achieve more than $80\%$ in IoU. When removing the visibility grid, we see a drop in mIoU from $36.3\%$ to $34.3\%$ as a result of losing this critical ambient space prior. 

\textbf{Voxel-based encoders are more suitable for completion than point-based encoders.} Rather than employing a 4D sparse convolution encoder, an alternative approach involves using a 4D point cloud encoder, such as P4Transformer \cite{fan2021point}, to directly leverage the point clouds. This substitution results in a $5.6\%$ decrease in IoU and a $6.8\%$ decrease in mIoU. This outcome substantiates the assertion that voxel-based methods are more effective for large-scale completion.. 

\section{Conclusion}
We introduce a new task of semantic complete scene forecasting (the SCSF task) from a 4D point cloud sequence and propose a new backbone SCSFNet for 4D scene understanding, equipped with hybrid geometric representations, attention-based skip connections, and visibility grids. To evaluate our method, we introduce IGPLAY and
IGNAV, two large-scale 4D egocentric vision datasets with complete geometry and semantic annotations covering abun-
dant indoor scenes and rich dynamics. Extensive experiments on our two high-quality indoor datasets and the outdoor SemanticKITTI benchmark not only confirm the significance of jointly modeling geometry forecasting and semantic completion but also demonstrate the effectiveness of our method.

\bibliography{main}

\newpage






\twocolumn[\begin{center}
    \LARGE\bfseries Semantic Complete Scene Forecasting from a 4D Dynamic Point Cloud Sequence Technical Appendix\par

    \vspace{1.5cm}
    \end{center}]

This document provides a list of supplemental materials to support the main paper.

\begin{itemize}
        \item {
            \textbf{Dataset Description. }We detail the generation process and statistics of IGPLAY and IGNAV in  Section \ref{iccv_dataset_desc}. Multi-type data formats allow us to conduct a variety of experiments related to the SCSF task.
        }

        \item {
            \textbf{Experiment Details. }We specify the computing infrastructure used for running experiments. In order to fit other relevant models to the new SCSF task compared with SCSFNet, we need to do some modifications. We will give details about this process in Section \ref{iccv_baseline_detail}. 
        }
        
        \item {
            \textbf{Additional Ablation Studies. }We provide additional
ablation studies in Section \ref{iccv_ablation_AAAI}. Specifically, we compare the performance of our 4D sparse voxel-based encoder with a point-based encoder and explore the impact of data augmentation on enhancing the performance of SCSFNet.
        }
        
        \item {
            \textbf{More results about Movable IoU. }We elaborate further on the calculation of Movable IoU for the SCSF task in Section \ref{iccv_mov}. The results in Section \ref{iccv_mov} confirm our claim that SCSFNet can better handle dynamic objects. 
        }

        \item {
            \textbf{More results about Semantic Scene Completion. }Ignoring future forecasting, we can degenerate  SCSFNet to tackle SSC (Semantic Scene Completion) problem for IGPLAY and NYUCAD in the form of 3D or 4D input in Section \ref{iccv_4DSSC} and get great performance.
        }
        
        \item {
            \textbf{Additional Visualizations. } We provide more visualizations in Section \ref{iccv_visual}, trying to explore more dynamic scenes for the SCSF task and prove the effectiveness of our SCSFNet. 
        }

        \item{
            \textbf{Limitations and Future work. } We analyze the limitations of our work based on the existing datasets experiments in Section \ref{limitations}. We hope that future work will try to solve these problems for the SCSF task.
        }
        
    \end{itemize}

\setcounter{section}{0}
\renewcommand{\thesection}{\Alph{section}}

\section{Dataset Description} \label{iccv_dataset_desc}

\subsection{iGibson}
iGibson \cite{igibson} is a simulation environment based on Bullet \cite{pybullet} that provides fast visual rendering and physical simulation. iGibson is equipped with fifteen fully interactive high-quality scenes, and hundreds of large 3D scenes reconstructed from real homes and offices, providing 12000+ additional interactive scenes. Some of the features of iGibson include domain randomization, integration with motion planners, and easy-to-use tools to collect human demonstrations. Based on the kinematic and extended states, iGibson 2.0 implements a list of logic predicates that indicate whether a logical state is true or false by checking conditions. The logic states implemented and checked by iGibson 2.0 such as ``inside of", ``on top of", ``next to", and so on. This system makes it more convenient to generate datasets.

\subsection{Dataset Generation}

To generate the simulated scenes in IGPLAY, 
we first specify the desired configuration of the environment in the logic language BDDL. In a typical BDDL description that we use, there is 1 floor, 2 tables, 1 bed, 2 sofas, 2 chairs, 3 floor lamps, 2 table lamps, and 1 iGibson robot. We then specify the desired logical state to make sure that everything is on the floor except table lamps which are on the table. After iGibson succeeds in creating the ideal scenario described by BDDL, we randomly discard some objects in the scene and then add two types of interactive and movable objects at random locations in the scene: toys and pops. These objects will drop on the table, on the sofa, on the chair, or just on the ground. This process was repeated and finally, we captured 1000 scenarios, each lasting 10 timesteps, with various data captured from the robot which is interacting with toys among the furniture.

While IGPLAY already exhibits many features suitable for 4D forecasting, the whole dynamics of the scene come from the viewer's movement and interaction with the objects. To provide a new perspective of the dynamics, we provide a new dataset called IGNAV with several robots navigating in the scene. The generation process of IGNAV is similar to that of IGPLAY. In order to make IGNAV different from IGPLAY, we made some adjustments in the semantic categories. 
For IGNAV, we similarly use the BDDL description to generate 1 floor, 2 tables, 1 bed, 2 chairs, 3 floor lamps, 2 table lamps, and 4 iGibson robots on average. Every robot will move at a certain velocity and direction. We captured 600 scenarios, each lasting 10 timesteps. Several robots move around the scene in their own way and interact with each other, with another robot watching them.

\subsection{Dataset statistics}

\begin{figure*}[t]
  \includegraphics[width=2\columnwidth]{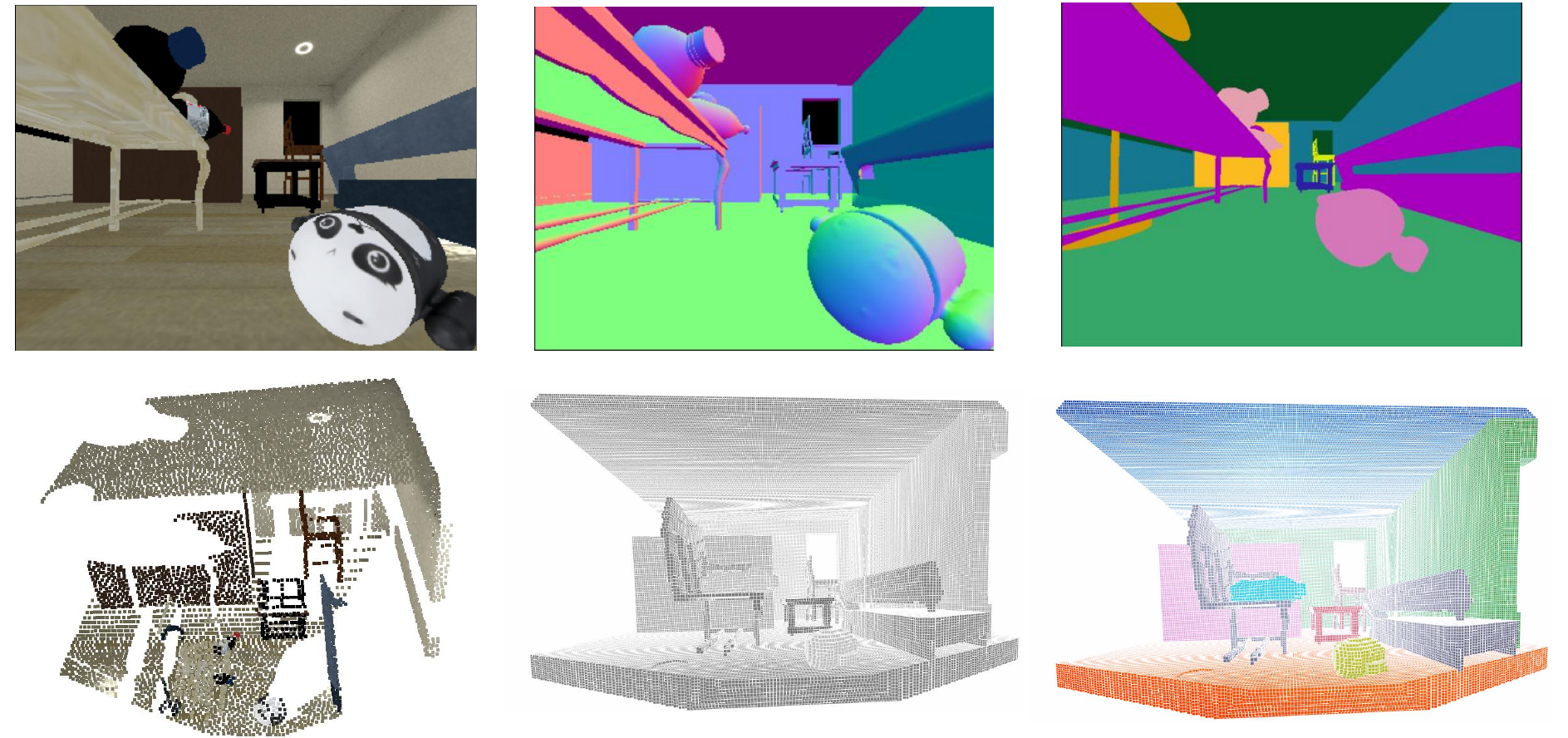}
  \caption{\textbf{Visualizations for IGPLAY.} We choose one frame in a sequence from IGPLAY. In the first row, we show the 2D RGB image, the 2D normal image, and the 2D semantic image. In the second row, we show the input 3D point cloud, the 3D voxel-based ground truth for the CSF task, and the 3D voxel-based ground truth for the SCSF task.}
\label{fig:igplay_igplay}
\end{figure*}  

We provide multiple data formats including RGB, normal, semantic labels, the point cloud, and the voxel-based semantic complete scene, which are shown in Figure \ref{fig:igplay_igplay}. 

Each scene in IGPLAY contains 10 semantic classes: ceiling, floor, wall, table, chair, bed, sofa, lamp, toy, and pop, with partial and complete occlusions happening as the viewer moves and interacts with objects. 

Each scene in IGNAV contains 9 semantic classes: ceiling, floor, wall, table, chair, sofa, lamp, window/door, and robot, with partial and complete occlusions happening as iGibson robots move and interact with each other or the furniture. 



In each timestep of our IGPLAY and IGNAV, we provide the 2D input image ($480 \times 640$) with RGB, normal and semantic labels, the 3D input point cloud ($32768$ points, Farthest Point Sampling from the point cloud generated by the 2D image), the 3D sparse voxel representation of the ground truth for the CSF task and SCSF task, and the visibility grid. The resolution of the ground truth is $240 \times 144 \times 240$ with grid size 0.02m, To store data more efficiently, we just record the non-empty voxels. 
For IGPLAY and IGNAV, the dimensions of the 3D
space is 4.8m horizontally, 2.88m vertically, and 4.8m in
depth. We also provide the intrinsic parameters of the iGibson camera and the field of view of the agent.
The train/test split is 80\%/20\%. We have shown typical data clips in the video we provide. 

\section{Experiment Details} \label{iccv_baseline_detail}

Our experimental results presented in the manuscript (like Table 1 and Table 2) use the same experimental configuration. Training a single SCSFNet model takes approximately 50 epochs in Ubuntu, with each epoch taking around 10 minutes. We utilize 2 Nvidia GeForce RTX 3090 GPUs, each with 24GB of VRAM. Our chosen optimizer is AdamW with a learning rate of 0.01, and we employ a batch size of 4. For streamlined management of our Python environment, we use Anaconda (version 1.9.0) and Vscode (version 1.75). And we use Python (version 3.8.16), Pytorch (1.8.1), MinkowskiEngine(0.5.4) and KNN-CUDA(0.2). Detailed instructions regarding our virtual Python environment will be provided once we release our code on GitHub in the near future.

\subsection{TLFPAD}
This baseline \cite{tlfpad} is tailored for processing a 4-frame sequence of point clouds to predict the next frame. Hence, we can adapt this baseline to perform the (surface) point cloud forecasting task with minimal modifications. However, due to its convolutional structure requiring 4-frame input, we replicate the first frame twice to create a 4-frame point cloud sequence. Each frame in the sequence comprises 8192 points.

As TLFPAD is not capable of managing occlusions and generating comprehensive future forecasts, our approach involves an initial step of forecasting a partial point cloud using TLFPAD. In this process, TLFPAD calculates motion vectors from the last input frame to generate the predicted frame, and we ensure that the point features such as color from the last frame are assigned to the predicted frame. Subsequently, employing the point cloud with features in the predicted frame, we proceed to perform semantic scene completion using the module from our SCSFNet.

We adhere to the default configuration of this baseline and conduct training for 100 epochs. Finally, we report the optimal outcome achieved on the validation set.
    
\subsection{Occlusion4D}
The baseline \cite{vanhoorick2022revealing} is designed for future scene completion. The input is a contiguous sequence of point cloud frames from a single view used to obtain an implicit representation.
    
In the training procedure, it samples solid points close to the target point cloud as positive instances and air points far away from the target point cloud as negative instances. Then it concatenates positive and negative instances to obtain query points and trains the network to predict properties on these points including density, semantic label and so on based on the implicit representation of the scene.

In the evaluation procedure, it doesn't have the target point cloud, so it just randomly samples query points and uses the network to predict the properties of these points. To obtain the predicted point cloud, it filters out query points with densities larger than a given threshold.

We need to modify it for the CSF and SCSF task.  For instance, We use contiguous sequences of 3 point cloud frames \(S_0,S_1,S_2\) with color as the input for IGPLAY. we convert the ground truth voxels in the next frame to a point cloud \(V\) and use farthest point sampling to sample 6144 points as our target point cloud. Due to the limitation of memory for this model, each frame contains 6144 points. We train the network to predict density on query points. In the evaluation procedure, we use  \(240\times 144\times 240\) query points, each of which represents a voxel with a side length of 0.02m. For those voxels with predicted densities larger than the density threshold, we assume they are occupied. Then we can calculate IoU for the CSF task.

\subsection{ST3DCNN}
This baseline \cite{mersch2022self} utilizes a novel 3D representation composed of concatenated range images as input. It simultaneously predicts future range images by an encoder-decoder structure and assigns per-point scores indicating validity or invalidity for multiple future time steps.
We refer to this approach as ST3DCNN, which we consider as the SOTA approach (without sensor extrinsics) on the KITTIOdometry dataset, in line with the description from \cite{khurana2023point}.

For SemanticKITTI, we maintain the same field of view (FOV) as in KITTIOdometry. However, we only need to forecast the point cloud in front of the vehicle. To achieve this, we adjust the range of yaw angles from [$-\pi$, $\pi$] to [$-\pi / 2$, $\pi / 2$]. Additionally, we resize the range map from $2048 \times 96$ to $1024 \times 96$. For IGPLAY and IGNAV, we modify the FOV to $1/3 \pi$ and resize the range map to $256 \times 256$.

It is noteworthy that the performance of ST3DCNN does not surpass that of TLFPAD. This discrepancy can be attributed to the fact that ST3DCNN's pipeline solely predicts the point cloud without incorporating additional information like color or semantic logits. In contrast, TLFPAD benefits from information derived from the last frame due to its scene-flow approach. If the ST3DCNN pipeline could accommodate supplementary aspects such as `color range map' or `semantic range map', it might yield enhanced results.

We stick to the default configuration of this baseline and train the model for 50 epochs. Subsequently, we select 8192 valid points from the predicted range map to generate the forecasted point cloud for the next frame. For semantic scene completion, we train a SCSFNet following the same approach as TLFPAD.

\section{Additional Ablation Studies} \label{iccv_ablation_AAAI}

\newcolumntype{Y}{>{\centering\arraybackslash}X}

\begin{table}[t]
\setlength{\tabcolsep}{3mm}
\centering
\footnotesize
\begin{tabularx}{1.0\linewidth}{lYY}
    \toprule
    \multicolumn{1}{c}{Methods} & \multicolumn{1}{c}{IoU} & \multicolumn{1}{c}{mIoU} \\
    \midrule
    No IF & 47.0 & 26.4  \\ %
    No  AS & 54.3 & 31.5  \\
    No VG & 55.7 & 34.3  \\
    P4T encoder \cite{fan2021point} & 50.9 & 29.5  \\
    Full Supervision  & 56.3   & 36.3 \\
    use DA   & 56.9 & 36.5 \\
    Ours & 56.5 & 36.3  \\ %
    \bottomrule
\end{tabularx}
\caption{\textbf{Ablation study of SCSFNet on IGPLAY for the SCSF task.} IF means implicit fields. AS means attention-based skip connections. VG means visibility grids. P4T means P4transformer. DA means data augmentation.}
\label{tables:ablations_iccv_supp}
\end{table}

\setlength{\tabcolsep}{2pt} 

We have discussed some components of SCSFNet including the implicit field (IF), attention-based skip connections (AS), visibility grids (VG), and the 4D sparse voxel-based encoder compared with point-based methods in the manuscript. Furthermore, we did more ablation studies on IGPLAY considering other factors, as shown in Table \ref{tables:ablations_iccv_supp}.

\textbf{Supervising implicit functions through on-the-fly random samples is both effective and efficient.} During the training phase, we employ random sampling of $10^5$ points from all the central coordinates of ground truth voxels. Another choice is using the entire ground truth, which entails calculating the loss based on $240 \times 144 \times 240 = 8294400$ voxels for a single data point in IGPLAY. This method, however, proves time-consuming and yields slightly inferior results (IoU drops from 56.5\% to 56.3\%). This substantiates that employing a random subset of $10^5$ voxels on the fly is significantly faster while maintaining effectiveness.


\textbf{Data augmentation improves the performance of SCSFNet.} 
Initially, we used a division scheme that selected $(0,1,2)$, $(3,4,5)$, and $(6,7,8)$ as input indices, with $(3)$, $(6)$, and $(9)$ as output indices for a 4D point cloud sequence of 10 frames in IGPLAY. This approach ensured no input overlap. However, to maximize data utilization, we adopted a more effective strategy by selecting $(0,1,2), (1,2,3),\ldots,(6,7,8)$ as input indices. This augmentation technique results in training the network three additional times for a single sequence in IGPLAY. This approach yielded a slight improvement, boosting IoU from 56.5\% to 56.9\% and mIoU from 36.3\% to 36.5\% for our SCSFNet.

\section{More results about Movable IoU} \label{iccv_mov}

\begin{table*}[t]
\setlength{\tabcolsep}{6pt}
\centering\small
\resizebox{\textwidth}{!}{
\begin{tabular}{|l|c c c c c c c c c c|c c c|} 
\hline
 &   \multicolumn{10}{c|}{Every Category}  & \multicolumn{3}{c|}{Avg.}\\ \hline
Methods   & ceil. & floor & wall & table & chair & bed & sofa & \textcolor{blue}{toy} & \textcolor{blue}{pop} & lamp  & \textcolor{blue}{IoU(M)} & IoU(S) & mIoU\\ 
\hline
 
TLFPAD  &  64.4 & 55.7 & 31.2 & 8.6 & 3.1 & 42.1 & 25.1 & 10.3 & 13.4 & 4.5 & 11.9 & 29.3 & 25.8 \\

ST3DCNN  &  65.4 & 62.9 & 32.9 & 9.5 & 4.3 & 39.0 & 19.2 & 6.7 & 11.9 & 3.2 & 9.3 & 29.6 & 25.5 \\

SCSFNet  &  70.7 & 65.0 & 37.9 & 20.7 & 12.0 & 53.4 & 32.8 & 24.2 & 32.1 & 14.6 &  \textbf{28.2} & \textbf{38.4} & \textbf{36.3} \\

\hline
\end{tabular}
}
\vspace{0.0cm}
\caption{\textbf{SCSF Results on dynamics of IGPLAY}. We list IoU of each specific category (10 classes). ``toy'' and ``pop'' in blue are movable classes. IoU(M) in  Avg. means Average Movable IoU. IoU(S) in Avg. means Average Static IoU.}
\label{tab:iccv_mov_igplay}
\end{table*}

\begin{table*}[t]
\setlength{\tabcolsep}{6pt}
\centering\small
\resizebox{\textwidth}{!}{
\begin{tabular}{|l|c c c c c c c c c|c c c|} 
\hline
 &   \multicolumn{9}{c|}{Every Category}  & \multicolumn{3}{c|}{Avg.}\\ \hline
Methods   & ceil. & floor & wall & table & chair & window & sofa   & \textcolor{blue}{robot} & lamp & \textcolor{blue}{IoU(M)} & IoU(S) & mIoU\\ 
\hline
 
TLFPAD  &  78.1 & 65.0 & 42.6 & 1.7 & 1.5 & 19.1 & 16.9 & 2.1 & 5.8 & 2.1 &  28.8 & 25.9 \\

ST3DCNN  &  74.2 & 70.1 & 51.6 & 8.4 & 7.1 & 12.5 & 7.1 & 0.0 & 3.5 & 0.0 & 29.3 & 26.1 \\

SCSFNet  &  83.6 & 72.5 & 58.5 & 14.8 & 24.2 & 37.4 & 35.8 & 18.3 & 14.0 & \textbf{18.3} & \textbf{42.6} &  \textbf{39.9} \\

\hline
\end{tabular}
}
\vspace{0.0cm}
\caption{\textbf{SCSF Results on dynamics of IGNAV}. We list IoU of each specific category (9 classes). ``robot'' in blue is the movable class.}
\label{tab:iccv_mov_ignav}
\end{table*}

\begin{table*}[hbt!]
\setlength{\tabcolsep}{3pt}
\centering\small
\resizebox{\textwidth}{!}{
\begin{tabular}{|l|c c c c c c c c c c c c c c c c c c c|c c c|} 
\hline
 &   \multicolumn{19}{c|}{Every Category}  & \multicolumn{3}{c|}{Avg.}\\ \hline
Methods   & \rotatebox{90}{\textcolor{blue}{car} } & \rotatebox{90}{\textcolor{black}{bicycle} } & \rotatebox{90}{\textcolor{black}{motorcycle} } & \rotatebox{90}{\textcolor{blue}{truck} } & \rotatebox{90}{\textcolor{blue}{other-vehicle} } & \rotatebox{90}{\textcolor{blue}{person} } & \rotatebox{90}{\textcolor{blue}{bicyclist} } & \rotatebox{90}{\textcolor{blue}{motorcyclist}} & \rotatebox{90}{road }  & \rotatebox{90}{parking } & \rotatebox{90}{sidewalk } & \rotatebox{90}{other-ground } & \rotatebox{90}{building } & \rotatebox{90}{fence } &  \rotatebox{90}{vegetation } & \rotatebox{90}{trunk } & \rotatebox{90}{terrain } & \rotatebox{90}{pole } & \rotatebox{90}{traffic-sign } & \rotatebox{90}{\textcolor{blue}{IoU(M)}} & \rotatebox{90}{IoU(S)} & \rotatebox{90}{mIoU}\\ 
\hline
 
TLFPAD  &  24.7 & 0.0 & 0.6 & 2.3 & 2.1 & 0.1 & 0.0 & 0.0 & 37.4 & 8.3 &  20.0 & 0.1 & 17.8 & 5.8 & 23.4 & 5.2 & 24.0 & 6.8 & 2.2 & 4.9 & 11.7 & 9.5 \\

ST3DCNN & 23.8 & 0.2 & 0.9 & 2.3 & 1.9 & 0.3 & 0.2 & 0.0 & 51.3 & 6.6 & 25.1 & 0.0 & 14.2 & 4.4 & 20.6 & 1.8 & 29.3 & 0.1 & 0.1 & 4.8 & 11.9 & 9.6 \\

SCSFNet  &  29.6 & 4.8 & 11.2 & 27.7 & 13.9 & 2.8 & 0.1 & 0.0 & 41.3 & 18.9 &  23.9 &  1.0 & 22.2 & 10.5 & 27.5 & 15.3 & 27.3 & 18.1 & 10.0 & \textbf{12.4} & \textbf{17.8} & \textbf{16.1}\\

\hline
\end{tabular}
}
\vspace{0.0cm}
\caption{\textbf{SCSF Results on dynamics of SemanticKITTI}. We list IoU of each specific category (19 classes). ``car'', ``truck'', ``other-vehicle'', ``person'', ``bicyclist'', and ``motorcyclist'' in blue are the movable classes.}
\label{tab:iccv_mov_semantic}
\end{table*}

In section 5.2 of the manuscript, we compared results between movable and static classes and reported them in Table 2 in the manuscript. A class is labeled as movable if
the corresponding objects can move, such as robots in IGNAV or cars in SemanticKITTI.
We present a comprehensive breakdown of IoU values for each category and offer an in-depth analysis of the dynamics within each dataset.

In the IGPLAY scenario, the agent's observations encompass the entire scene, including interactions with movable toys and objects popping up from the ground. Hence, accurate forecasting on IGPLAY necessitates special attention to these elements. As shown in Table \ref{tab:iccv_mov_igplay}, our SCSFNet significantly outperforms TLFPAD in forecasting the semantic complete scene for dynamic objects like ``toy'' and ``pop''. Specifically, our SCSFNet demonstrates a remarkable 16.3\% enhancement in Movable IoU and a notable 9.1\% increase in Static IoU compared to TLFPAD.

In the IGNAV scenario, the mobility of robots adds complexity to predicting their trajectories, interactions, and occlusions. As depicted in Table \ref{tab:iccv_mov_ignav}, TLFPAD faces challenges, resulting in a significant 16.2\% reduction in Movable IoU and a noteworthy 13.8\% decrease in Static IoU.

For SemanticKITTI, classes such as ``car'' that encompass moving objects with annotations like ``moving-car'' from SemanticKITTI are considered movable classes. The results in Table \ref{tab:iccv_mov_semantic} highlight that SCSFNet yields notable enhancements, with a 7.5\% boost in Movable IoU and a commendable 6.1\% improvement in Static IoU.

Across all these outcomes, a clear pattern emerges: in contrast to static objects, SCSFNet excels in addressing dynamic objects, a capability that holds paramount significance for accurate future forecasting.

\section{Semantic Scene Completion tasks for SCSFNet} \label{iccv_4DSSC}

\begin{table*}[t]
\setlength{\tabcolsep}{3.6pt}
\centering\small
\resizebox{\textwidth}{!}{
\begin{tabular}{|l|c c c|c c c c c c c c c c|c|} 
\hline
 &  \multicolumn{3}{c|}{Scene Completion (SC)} & \multicolumn{11}{c|}{Semantic Scene Completion (SSC)} \\ \hline
Methods   & prec. & recall & IoU & ceil. & floor & wall & table & chair & bed & sofa & toy & pop & lamp  & avg. \\ 
\hline

SCSFNet(3D)  & 88.7 & \textbf{66.3} & 61.1 & \textbf{68.3} & 51.2 & \textbf{64.5} & 30.5 & \textbf{21.2} & 54.9 & 42.2 & 27.7 & 41.1 & 26.9 &  42.8 \\

SCSFNet(4D)  & \textbf{90.7} & 65.4 & \textbf{61.3} & 64.9 & \textbf{52.8} & 63.4 & \textbf{32.0} & \textbf{21.2} & \textbf{60.3} & \textbf{42.7} & \textbf{42.4} & \textbf{45.6} & \textbf{27.5} &  \textbf{45.3} \\

\hline
\end{tabular}
}
\vspace{-0.0cm}
\caption{\textbf{SSC Results on IGPLAY dataset}. We show the results of SCSFNet in 3D and SCSFNet in 4D for SSC problem. For better comparison such as movable objects, we list IoU of each specific category.}
\label{tab:4dssc}
\vspace{-0.1cm}
\end{table*}

\begin{table*}[t]
\setlength{\tabcolsep}{3.6pt}
\centering\small
\resizebox{\textwidth}{!}{
\begin{tabular}{|l| c c c|c c c c c c c c c c c | c|} 
\hline
 &  \multicolumn{3}{c|}{Scene Completion (SC)} & \multicolumn{12}{c|}{Semantic Scene Completion (SSC)} \\ \hline
Methods   & prec. & recall & IoU & ceil. & floor & wall & win. & chair & bed & sofa & table & tvs & furn. & objs. & avg. \\ 
\hline
SSCNet~\cite{song2017semantic}  & 75.4 & \textbf{{96.3}} & 73.2 & 32.5 & 92.6 & 40.2 &  8.9 & 33.9 & 57.0 & 59.5 & 28.3 &  8.1 & 44.8 & 25.1 & 40.0\\ 
DDRNet~\cite{li2019rgbd} 	& 88.7 & 88.5 & 79.4 & 54.1 & 91.5 & 56.4 & 14.9 & 37.0 & 55.7 & 51.0 & 28.8 & 9.2 & 44.1 & 27.8 & 42.8 \\
TS3D~\cite{garbade2019two}   & - & - & 76.1 & 25.9 & 93.8 & 48.9 & 33.4 & 31.2 & 66.1 & 56.4 & 31.6 & 38.5 & 51.4 & 30.8 & 46.2 \\ 
CCPNet~\cite{zhang2019cascaded}  & 91.3 & 92.6 & 82.4 & 56.2 & \textbf{{94.6}} & 58.7 &\textbf{ 35.1} & 44.8 & 68.6 & 65.3 & 37.6 & 35.5 & 53.1 & 35.2 & 53.2 \\
Sketch~\cite{chen20203d}  & 90.6 & 92.2 & \textbf{84.2} & 59.7 & 94.3 & 64.3 & 32.6 & 51.7 & 72.0 & \textbf{{68.7}} & \textbf{45.9} & 19.0 & \textbf{60.5} & \textbf{38.5} &  55.2 \\ 
\hline
Ours (SCSFNet) & 
{\textbf{{92.8}}}& {89.1}& {{83.3}} & {\textbf{{60.6}}}& {94.3}& {\textbf{{65.5}}}& {{32.7}}& {\textbf{{55.3}}}& {\textbf{{74.5}}}& {{66.7}}& {{43.3}}& {\textbf{{42.4}}}& {{58.5}}& {\textbf{{38.5}}}& {\textbf{{57.5}}}\\
\hline
\end{tabular}
}
\vspace{0.0cm}
\caption{\textbf{SSC Results on NYUCAD dataset \cite{silberman2012indoor}}. We compare SCSFNet with models using the same straightforward pipeline and ignore models using iterative pipeline. }
\label{tab:NYUCAD}
\end{table*}

\begin{figure*}[hbt!]
  \includegraphics[width=2\columnwidth]{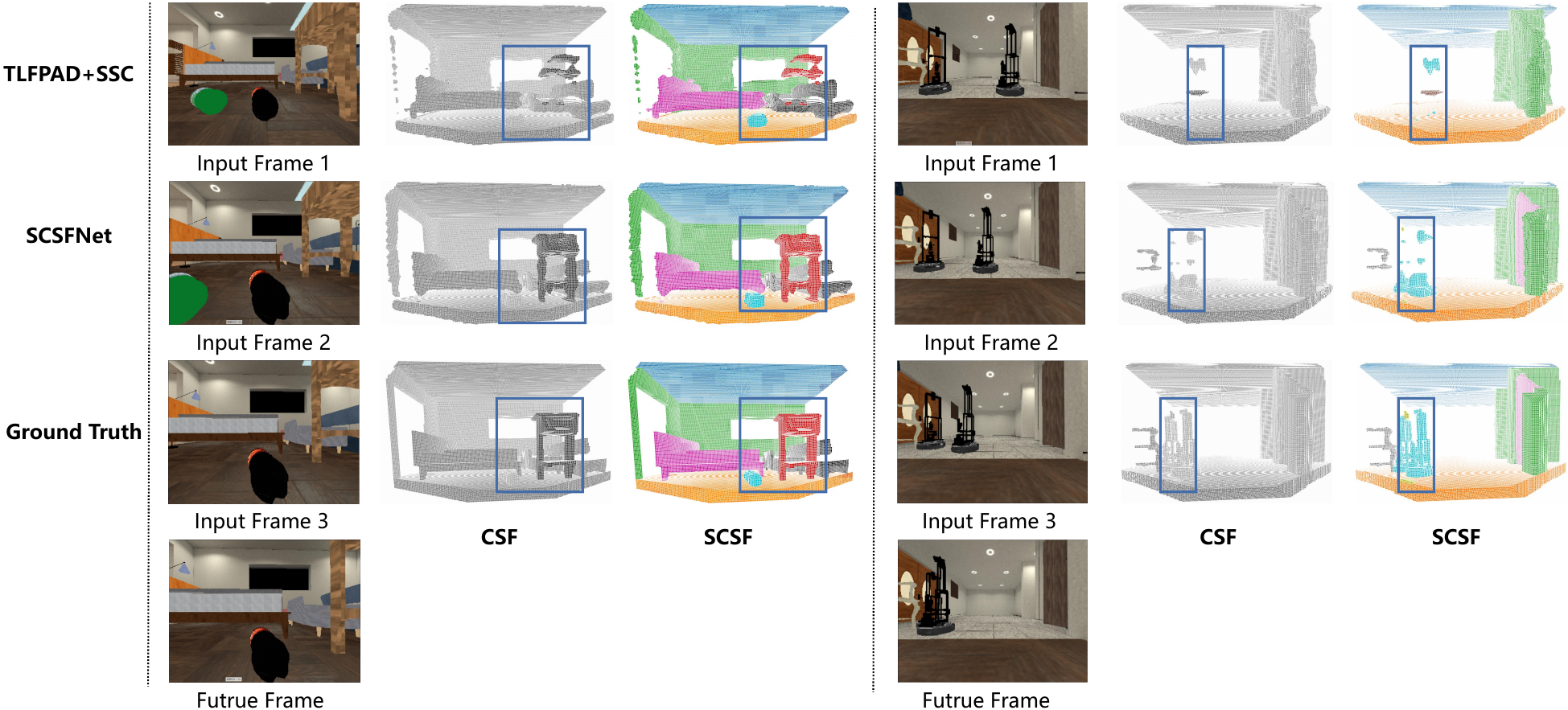}
  \caption{\textbf{Visualizations for our synthetic indoor datasets.} The sequence on the left is taken from IGPLAY. The agent saw a bottle of soda and a bottle of cola at first and approached the bottle of cola beside the table. The sequence on the right is taken from IGNAV. The agent is watching a robot going toward another stationary robot. SCSFNet can predict the shape of the bottle, the table, and the robots better under the change of perspective and dynamic objects.}
\label{fig:igibson_vis}
\end{figure*}   


Our paper proposed the SCSF task, which mainly focuses on future forecasting. If we do not forecast and only focus on the current frame (the last given frame), degenerate cases for CSF (Complete Scene Forecasting) and SCSF (Semantic Complete Scene Forecasting) become SC (Scene Completion) and SSC (Semantic Scene Completion). Our SCSFNet can solve the SSC problem using the same pipeline for the SCSF task in different dimensions.

\textbf{3D SSC.} We treat every frame in a 4D point cloud sequence as a single point cloud input ignoring the connections between different frames. In this way, we only feed one frame into the model and supervise this single frame.

\textbf{4D SSC.} We consider that connections between different frames in a 4D point cloud sequence are important to gain more information in scene understanding and completion, and our SCSFNet is capable of processing multiple frames simultaneously. In this way, we feed 3 consecutive frames into the model, generate three high-resolution voxel grids to represent the completion results, and then supervise these 3 frames simultaneously. We need a little modification of the network architecture in this setting.

The results in IGPLAY are shown in Table \ref{tab:4dssc}. For the SSC problem, we mainly use IoU and mIoU to evaluate the completion with semantic labels. SCSFNet (4D) gives a 0.2\% improvement in IoU and a 2.5\% improvement in mIoU compared with SCSFNet (3D). 

In particular, SCSFNet (4D) has an obvious forecasting advantage in the ``toy'' class. In IGPLAY, the agent interacts with these objects, thus they have strong motion continuity between successive frames. If multiple frames are taken together as the input, SCSFNet can model them jointly to gain more insights into the scene dynamics.

Similarly, we try our SCSFNet (3D) in NYUCAD\cite{silberman2012indoor}, one of the most famous SSC datasets. We compare our model to other state-of-the-art SSC methods using the same iterative pipeline (without additional instance labels) in Table \ref{tab:NYUCAD}. We can find that our SCSFNet is a generalizable model not only for future forecasting but also for classic semantic completion.

\section{Additional Visualizations} \label{iccv_visual}   


\begin{figure*}[hbt!]
  \includegraphics[width=2\columnwidth]{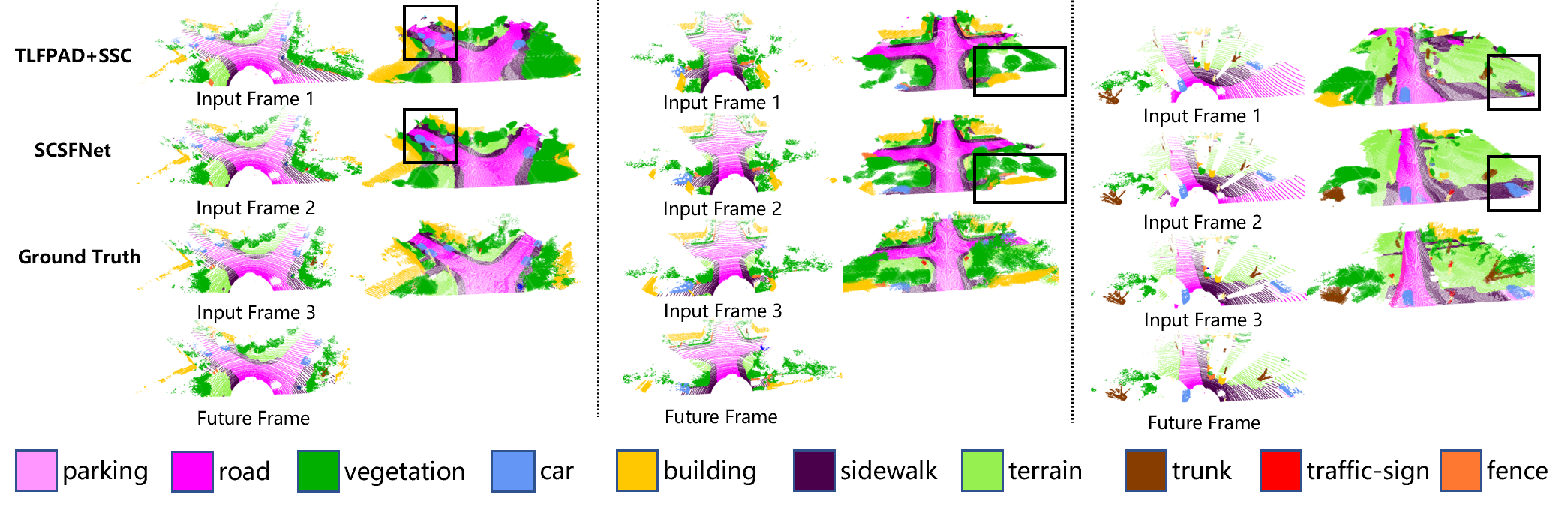}
  \caption{\textbf{Visualizations for SemanticKITTI.} We provide three groups of visualizations. Each group consists of two columns: the first column represents the three past frames of point clouds and the future frame, and the second column signifies SCSF predictions by different methods. We do not show the results for the CSF task because it's hard to distinguish objects without semantics. Although SemanticKITTI only provides raw point cloud data, we can use the semantic segmentation logits predicted by 2DPASS \cite{yan20222dpass}.}
\label{fig:semantickitti_vis}
\end{figure*}


We also offer enhanced visualizations of both the CSF task and the SCSF task across three distinct datasets. With IGPLAY and IGNAV shown in Figure \ref{fig:igibson_vis}, our aim is to present a more dynamic visualization and showcase the capability of  SCSFNet for semantic complete forecasting. As for SemanticKITTI \cite{behley2019iccv} shown in Figure \ref{fig:semantickitti_vis}, we strive to demonstrate the versatility of our SCSFNet, emphasizing its applicability in real-world scenarios.

Our datasets and detailed instructions for access will be made available to the public in the near future. And we will also provide detailed code for data visualization using the raw datasets IGPLAY and IGNAV.
\section{Limitations and Future work} \label{limitations}

In our work, we require dynamic scenes that can support completion
and forecasting at the same time. Given the limited options
for real-world datasets meeting these criteria, we find SemanticKITTI to be the only suitable choice, which consists
of large-scale 4D point cloud sequences with completion
and semantic annotations. To further demonstrate the effectiveness of our method in
the indoor setting, we generate 2 large-scale indoor datasets.
We hope to extend the evaluation
of our approach to suitable real-world datasets released in
the future.

Additionally, in our experiments, we only performed relatively short-term forecasting due to the inherent challenges of the SCSF task. This limitation hinders the model's ability to provide an accurate understanding of the entire scene, particularly focusing on dynamic objects, in the long term. In the future, we aspire to see more efficient approaches that can handle long-term SCSF tasks with minimal computational overhead.

\end{document}